\documentclass[review, 12pt, numbers, sort&compress]{elsarticle}

\usepackage{changes}

\usepackage{amsmath}
\usepackage{wasysym}
\usepackage{amssymb}  
\usepackage{amsfonts}  
\usepackage{multirow}
\usepackage{txfonts}  

\usepackage{textcomp}
\usepackage{stfloats}
\usepackage{url}
\usepackage{verbatim}
\usepackage{graphicx}
\usepackage{dsfont}
\usepackage{multirow}
\usepackage{color}
\usepackage{caption}  

\usepackage{geometry}
\usepackage[font=small,labelfont=bf,labelsep=none]{caption}  
\usepackage{balance}

\usepackage{setspace}
\usepackage[T1]{fontenc} 
\usepackage{booktabs}     
\usepackage{algorithm}  
\usepackage{algorithmic}   
\usepackage{float}
\usepackage{epstopdf}
\usepackage{subfigure}
\RequirePackage{caption}



\usepackage{siunitx}
\usepackage{bm}
\usepackage{array}  
\makeatletter
\newcommand\multiline[1]{\parbox[t]{\dimexpr\linewidth-\ALG@thistlm}{#1}}
\makeatother
\usepackage{calligra}

\PassOptionsToPackage{noend}{algpseudocode}
\def\dif{\mathop{}\hphantom{\mskip-\thinmuskip}\mathrm{d}}%
\let\daccent\d
\let\d\relax
\newcommand\d{\ifmmode\dif\else\expandafter\daccent\fi}

\usepackage[english]{babel}   
\allowdisplaybreaks[4]   
\usepackage{etoolbox}
\makeatletter
\patchcmd{\@makecaption}
{\scshape}
{}
{}
{}
\makeatletter
\patchcmd{\@makecaption}
{\\}
{.\ }
{}
{}
\makeatother

\errorcontextlines\maxdimen

\journal{Signal Processing}

\begin{document}
	
\begin{frontmatter}  

\title{IMMNet: Hybrid Fusion of Model-based and Data-driven Approaches for Maneuvering Target Tracking}

\author[SEU]{Yixuan Zhao}
\ead{zhaoyixuan123321@126.com}

\author[SEU]{Chaoqun Yang\corref{cor1}}
\ead{ycq@seu.edu.cn}
\cortext[cor1]{Corresponding Author}

\author[UES]{Lin Gao}
\ead{lingao_1014@126.com}

\author[SEP]{Yongxiao Tian}
\ead{tianyongxiao1991@163.com}

\author[SJTU]{Ting Yuan}
\ead{tyuan@sjtu.edu.cn}

\address[SEU]{School of Automation, Southeast University, Nanjing, 210096, China}
\address[UES]{School of ICE, University of Electronic Science and Technology of China, Chengdu, 611731, China}
\address[SEP]{Faculty of Artificial Intelligence, Shanghai University of Electric Power, Shanghai, 201300, China}
\address[SJTU]{School of Automation and
Intelligent Sensing, Shanghai Jiao Tong University, Shanghai, 200240,
China}
\begin{abstract}
Maneuvering target tracking in three-dimensional space remains a challenging problem due to complex motion dynamics and model mismatch. To address this, this paper proposes
a hybrid model/data-driven algorithm named IMMNet, which  integrates the interpretable structure of the interacting multiple model (IMM) algorithm with learnable neural components. Unlike end-to-end black-box methods, the proposed IMMNet algorithm not only can preserve the Bayesian  inference mechanism that is  essential for real-time radar applications, but also can  adaptively learn motion patterns and noise characteristics from data. 
Extensive experiments demonstrate that the proposed IMMNet algorithm consistently outperforms the existing algorithms across various scenarios, validating it as a robust, interpretable, and practical solution for maneuvering target tracking.
\end{abstract}

\end{frontmatter}

\section{Introduction}
Recently, maneuvering target tracking (MTT) has become increasingly essential with its widespread applications in radar surveillance, autonomous vehicles and missile defense~\cite{ZHOU2026110285, DU2025109731, 10689476}. However, MTT in complex dynamic environments poses significant challenges due to highly variable motion modes and unpredictable trajectory characteristics of targets~\cite{XIA2026110322, Bai2025Radar, 10726762}. Unlike those targets with steady-state or slowly varying dynamics, maneuvering targets frequently switch between different motion modes, leading to rapid changes in velocity and acceleration~\cite{garcia2015model, nadarajah2012imm}. Such behaviors make it difficult for traditional estimation algorithms to maintain stable and accurate tracking~\cite{jia2017highly}, thereby highlighting the critical need for new state estimation algorithms.

To address the challenges of MTT, various algorithms of state estimation have been developed such as  Singer maneuvering model~\cite{singer2007estimating}, input estimation method~\cite{chan2007kalman} and  interacting multiple model (IMM) algorithm~\cite{blom2002interacting}, etc. Among them, the IMM algorithm stands out as one of the most classical and effective solutions. To cope with time-varying target dynamics, it models target motion as a switching Markov process and fuses state estimates from multiple parallel motion models. However, the above algorithms are model-driven, which means that they depend critically on precise state-space formulations and accurate estimation of process and measurement noise covariance matrices. In practical scenarios involving model mismatch, parameter drift, or environmental disturbances, their performance may degrade significantly. 

To overcome the inherent limitations of the above model-driven algorithms, researchers have increasingly turned to data-driven algorithms. For example, Liu et al. employed neural networks to generate correction terms for estimates of classical filters for MTT~\cite{liu2020deepmtt}. Gao et al. utilized RNNs to learn target kinematics directly from raw radar sequences without any predefined motion models~\cite{gao2018maneuvering}. Zhang et al. proposed an encoder-decoder Transformer architecture named TrMTT~\cite{zhang2024deep}. Data-driven algorithms exhibit promising advantages, particularly strong adaptability to unknown noise and model mismatch. However, several critical challenges still persist in their practical applications. First, most existing data-driven algorithms suffer from limited interpretability, excessive parameter quantities, and dependence on fixed-length observation windows, which restricts their applicability in real-time applications. Furthermore, data-driven algorithms usually determine model probabilities through black-box operations, thereby reducing their consistency with rigorous Bayesian inference principles.

Accordingly, a natural question arises as to whether we can integrate data-driven MTT algorithms and model-driven MTT algorithms to develop a new MTT algorithm that inherits the advantages of them. Therefore, this paper focuses on the design of the hybrid-driven MTT algorithm. We propose IMMNet, a hybrid model/data-driven algorithm specifically designed for MTT. We integrate the principled, interpretable structure of the IMM framework with the adaptive learning capabilities of neural networks, which forms an interpretable hybrid-driven MTT algorithm. In the IMMNet algorithm, we first leverage the Transformer architecture~\cite{NIPS2017_3f5ee243} to replace the fixed model probability update mechanism of the IMM algorithm, enabling the capture of complex spatiotemporal dependencies without the need of Markovian assumptions. 
Then, we incorporate a hybrid model/data-driven filter named KalmanNet~\cite{revach2022kalmannet}, which replaces the analytical Kalman gain computation with a neural module to adapt to unknown noise statistics and mitigate model mismatch through end-to-end trainable parameter learning.
In this way, the proposed IMMNet algorithm combines the strengths of model-driven and data-driven algorithms. The model-driven algorithms contribute interpretability and a principled structure, while the data-driven algorithms provide adaptability and robustness, allowing the proposed IMMNet algorithm to inherited their merits of both. The main contributions of this paper are summarized as follows:

\begin{enumerate}

\item We propose IMMNet, a novel hybrid model/data-driven algorithm for MTT which integrates KalmanNet-based state estimation with a Transformer-based motion-mode classifier. By replacing the fixed Markov transition and traditional Kalman filters in the IMM algorithm with learnable neural modules, the proposed IMMNet algorithm achieves stronger tracking adaptability for complex 3D maneuvering targets and effectively alleviates performance degradation caused by model mismatch.


\item We build a large-scale, physically realistic dataset for 3D MTT, which supports multi-mode-switching within trajectories, covers realistic radar surveillance ranges and aircraft kinematics, and provides synchronized state, measurement, and motion-mode labels.

\end{enumerate}

The remainder of this paper is organized as follows. Section~\ref{work} reviews related work. Section~\ref{system model} introduces preliminaries. Section~\ref{system} presents system model. Section~\ref{immnet} details the architecture of the proposed IMMNet algorithm. Section~\ref{dataset} describes  the dataset of 3D MTT. Section~\ref{EXPERIMENTAL} provides simulation experiments. Finally, section~\ref{conclusion} concludes the paper.

\section{Related Work}~\label{work}
Numerous studies have been dedicated to MTT so far. According to the core tracking mechanisms, most of the existing MTT algorithms can be divided into two categories: model-driven algorithms and data-driven algorithms. 

For the model-driven algorithms, besides the model-driven algorithms mentioned in section~1, variety of motion models and filtering techniques have been developed to handle MTT. 
For example, Dong et al. proposed an expected mode augmentation-based variable structure multiple model GMCPHD filter with adaptive model sets and gating for maneuvering multi-target tracking.~\cite{dong2017maneuvering}.
Zhou et al. proposed a switch-constrained multiple-model algorithm and a decision-aided version, which restricts model switching within three consecutive steps and uses likelihood-based detection to suppress peak errors~\cite{zhou2023switch}.
Matei et al. proposed improved IMM estimators with a nonzero-mean white noise turn-rate model for tracking sharply maneuvering ground targets \cite{visina2018multiple}.
However, most of these model-driven MTT algorithms rely heavily on accurate motion models and noise covariance parameters, leading to performance degradation under model mismatch or uncertain environments.

For the data-driven algorithms, besides  DeepMTT ~\cite{liu2020deepmtt}, RNN based trackers \cite{gao2018maneuvering}, and transformer based TrMTT ~\cite{zhang2024deep}, more specialized networks have been explored. 
Considering the joint problem of tracking and classification, Yu et al. proposed a deep learning algorithm to estimate states and identify motion modes of maneuvering targets ~\cite{yu2022deep}. 
Yu et al. proposed DeepGTT, a general deep learning tracking algorithm that learns target dynamic laws via LSTM and achieves trajectory tracking without predefined motion models~\cite{yu2021deepgtt}.
Chen et al. proposed a data-driven intelligent multiframe joint tracking method for maneuvering targets in clutter environments, which integrates adaptive maneuver-aware sampling, pyramid spatio-temporal graph construction and interactive Transformer-graph attention autoencoder\cite{chen2025data}.
Nevertheless, most of the above data-driven MTT algorithms suffer from low interpretability, large parameter scales, and reliance on fixed-length observation windows, limiting their real-world deployment.


Although several hybrid model/data-driven algorithms have been proposed for MTT, related research still remains limited.
In ~\cite{han2023multi}, a multi-model KalmanNet is introduced, which uses a dynamic routing network to adaptively weight predefined motion models. 
In summary, although the above work has advanced the study of MTT significantly, there are still some limitations. The existing model-driven algorithms depend on precise state space formulations and accurate estimation of noise covariance matrices, which makes them sensitive to model mismatch. The existing data-driven algorithms suffer from limited interpretability, excessive parameter quantities, and a dependence on fixed length observation windows. The existing hybrid model/data-driven algorithms fail to achieve deep fusion of data-driven and model-driven mechanisms and are limited by improper dataset settings, resulting in unsatisfactory tracking performance. Thus, hybrid model/data-driven MTT algorithms still deserves further in-depth research.

\section{Preliminaries}~\label{system model}
As preliminaries, this section briefly reviews the KalmanNet and IMM algorithms.
\subsection{KalmanNet Algorithm}

\begin{figure}[h]
	\centering
	\includegraphics[width=0.5\textwidth]{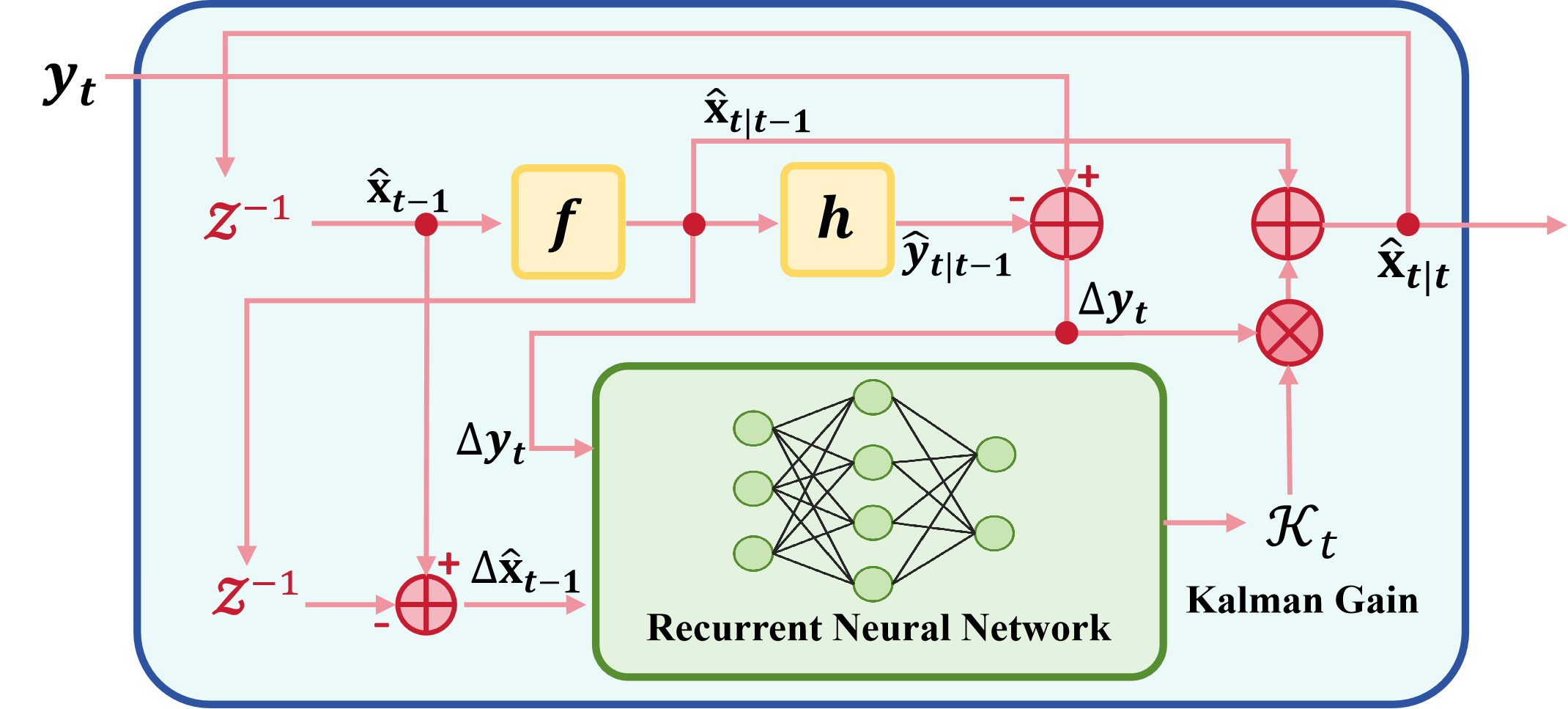}\\
	\caption{The architecture of KalmanNet.}\label{fig:kalmanet_arch}
\end{figure}

The KalmanNet algorithm is an interpretable, low-complexity, and data-efficient deep neural network-based state estimator ~\cite{revach2022kalmannet}. It integrates a neural network into the Kalman filter to learn Kalman gain directly from data, thereby eliminating the need for prior knowledge of process and measurement noise statistics. The overall architecture of the KalmanNet algorithm is illustrated in 
Fig.~\ref{fig:kalmanet_arch}.

KalmanNet operates in two steps: prediction and update. In the predicted step, given the posterior estimate $ \boldsymbol{x}_{t-1|t-1}$ from the previous time step $t-1$, it first computes the prior state estimate and the predicted measurement using the known system dynamics:
\begin{align}
    \boldsymbol{x}_{t|t-1} &= f(\boldsymbol{x}_{t-1|t-1}), \label{eq:pred_state} \\
    \boldsymbol{y}_{t|t-1} &= h(\boldsymbol{x}_{t|t-1}), \label{eq:pred_obs}
\end{align}
where $f(\cdot)$ and $h(\cdot)$ denote the state transition and measurement functions, respectively.
In the update step, the posterior state $\boldsymbol{x}_{t|t}$ estimate is obtained via
\begin{equation}
    \boldsymbol{x}_{t|t} = \boldsymbol{x}_{t|t-1} + \boldsymbol{K}_t (\boldsymbol{y}_t - \boldsymbol{y}_{t|t-1}), \label{eq:posterior}
\end{equation}
where $\boldsymbol{K}_t$ is the Kalman gain.
The key distinction lies in how $\boldsymbol{K}_t$ is obtained: Rather than being computed from analytical error covariance propagation, the KalmanNet algorithm employs a dedicated RNN to infer $\boldsymbol{K}_t$ directly from sequential input features derived from past measurements and innovations.

\subsection{IMM Algorithm}
The IMM algorithm is one of the most representative model-based algorithms for MTT. This algorithm executes multiple dynamic models simultaneously, and each model characterizes a specific motion mode. It integrates the estimation results through the real-time updated model probabilities, thereby accomplishing precise state estimation.

The IMM algorithm assumes that the target’s motion can be described by a finite set of dynamic models $\mathcal{M} = \{m^{(1)}, m^{(2)}, \dots, m^{(M)}\}$. Each model $m^{(i)}$ corresponds to a specific motion hypothesis and is typically represented by a state-space model. Associated with each model is a time-varying probability $\mu_t^{(i)}$, which quantifies the likelihood that model $m^{(i)}$ best describes the target’s current behavior. The IMM algorithm proceeds through the following steps at each time step $t$:

\textbf{1) Model set selection and initialization:}
Each model $ m^{(i)} $ is initialized with a prior state estimate $\boldsymbol{x}_{0|0}^{(i)}$ and an initial probability $\mu_{0}^{(i)}$, satisfying $\sum_{i \in \mathcal{M}}^{\mathcal{M}} \mu_0^{(i)} = 1$.

\textbf{2) Prediction:}  
For each model $ m^{(i)}$, the predicted state and covariance are computed as:
\begin{align}
    \boldsymbol{\hat{x}}_{t|t-1}^{(i)} &= \boldsymbol{F}^{(i)} \boldsymbol{\hat{x}}_{t-1|t-1}^{(i)}, \label{eq:imm_pred_state} \\
    \boldsymbol{P}_{t|t-1}^{(i)} &= \boldsymbol{F}^{(i)} \boldsymbol{P}_{t-1|t-1}^{(i)} (\boldsymbol{F}^{(i)})^\top + \boldsymbol{Q}^{(i)}, \label{eq:imm_pred_cov}
\end{align}
where $\boldsymbol{F}^{(i)}$ is the state transition matrix and $\boldsymbol{Q}^{(i)}$ is the process noise covariance for model $m^{(i)}$.

\textbf{3) Update:}  
Given measurement $\boldsymbol{z}_t$, each model performs a standard Kalman update:
\begin{align}
    \boldsymbol{K}_t^{(i)} &= \boldsymbol{P}_{t|t-1}^{(i)} (\boldsymbol{H}^{(i)})^\top \left( \boldsymbol{H}^{(i)} \boldsymbol{P}_{t|t-1}^{(i)} (\boldsymbol{H}^{(i)})^\top + \boldsymbol{R}^{(i)} \right)^{-1}, \label{eq:imm_kalman_gain} \\
    \boldsymbol{x}_{t|t}^{(i)} &= \boldsymbol{x}_{t|t-1}^{(i)} + \boldsymbol{K}_t^{(i)} \left( \boldsymbol{z}_t - \boldsymbol{H}^{(i)} \boldsymbol{x}_{t|t-1}^{(i)} \right), \label{eq:imm_updated_state} \\
    \boldsymbol{P}_{t|t}^{(i)} &= \left( \boldsymbol{I} - \boldsymbol{K}_t^{(i)} \boldsymbol{H}^{(i)} \right) \boldsymbol{P}_{t|t-1}^{(i)}, \label{eq:imm_updated_cov}
\end{align}
where $\boldsymbol{x}_{t|t}^{(i)}$ is the posterior state of the $i$-th Kalman filter, $\boldsymbol{K}_t^{(i)}$ is the Kalman gain, $\boldsymbol{H}^{(i)}$ is the observation matrix, $\boldsymbol{R}^{(i)}$ is the observation noise covariance.

\textbf{4) Model probability updating:}  
Using Bayes’ theorem, the updated model probability is:
\begin{equation}
    \mu_{t|t}^{(i)} = \frac{ \mathcal{N}\left( \boldsymbol{z}_t; \boldsymbol{H}^{(i)} \boldsymbol{x}_{t|t-1}^{(i)}, \boldsymbol{S}_t^{(i)} \right) \, \mu_{t|t-1}^{(i)} }{ \sum_{j=1}^{M} \mathcal{N}\left( \boldsymbol{z}_t; \boldsymbol{H}^{(j)} \boldsymbol{x}_{t|t-1}^{(j)}, \boldsymbol{S}_t^{(j)} \right) \, \mu_{t|t-1}^{(j)} }, \label{eq:imm_model_prob}
\end{equation}
where $\boldsymbol{S}_t^{(i)} = \boldsymbol{H}^{(i)} \boldsymbol{P}_{t|t-1}^{(i)} (\boldsymbol{H}^{(i)})^\top + \boldsymbol{R}^{(i)}$ is the innovation covariance, and $\mathcal{N}(\cdot; \boldsymbol{\mu}, \boldsymbol{\Sigma})$ denotes the Gaussian distribution with mean $ \boldsymbol{\mu} $ and covariance $ \boldsymbol{\Sigma} $.

\textbf{5) Interaction and fusion:}  
The final fused estimated state is computed as:
\begin{equation}
    \boldsymbol{x}_{t|t} = \sum_{i \in \mathcal{M}}^{\mathcal{M}} \mu_{t|t}^{(i)} \, \boldsymbol{x}_{t|t}^{(i)},
    \label{eq:imm_fused_state}
\end{equation}
which yields a robust estimate that adapts to maneuvering behavior over time.

\section{System Model}~\label{system}


\begin{figure}[h]
	\centering
	\includegraphics[width=0.5\textwidth]{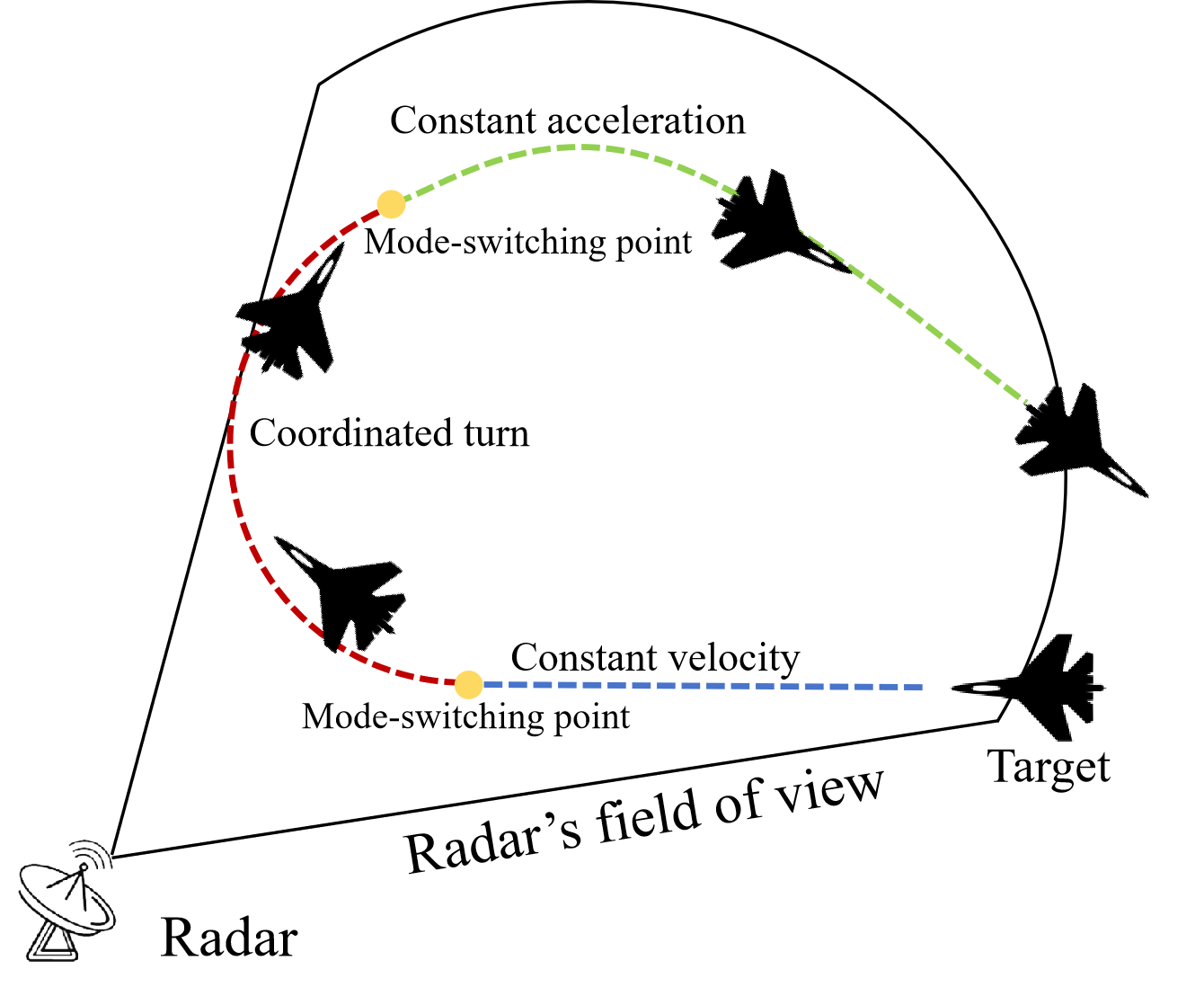}\\
	\caption{The considered MTT scenario.}\label{fig:maneuveringtar}
\end{figure}

\subsection{Dynamic Model}
As shown in Fig.~\ref{fig:maneuveringtar}, consider a maneuvering target whose motion model switches among different patterns. In this paper, the following four representative models are considered: constant velocity (CV), constant acceleration (CA), coordinated turn with maximum positive turn rate (CT+), and coordinated turn with minimum negative turn rate (CT-). It is worth noting that, the proposed algorithm in this paper is not restricted to the above four models. The selection of diverse maneuvering models can be tailored to specific application scenarios.

To fully represent 3D kinematics, the state of this target includes position, velocity, and acceleration along each spatial axis, resulting in a 9-dimensional state:
\begin{equation}
    \boldsymbol{x}_t^{(i)} = \left[ l_x,\, v_x,\, a_x,\, l_y,\, v_y,\, a_y,\, l_z,\, v_z,\, a_z \right]^\top,
    \label{eq:state_vector}
\end{equation}
where $l$, $v$, and $a$ denote position, velocity, and acceleration, respectively. The above state evolves according to the following linear Gaussian dynamics:
\begin{equation}
    \boldsymbol{x}_t^{(i)} = \boldsymbol{F}^{(i)} \boldsymbol{x}_{t-1}^{(i)} + \boldsymbol{w}_t, \quad \boldsymbol{w}_t \sim \mathcal{N}(\boldsymbol{0}, \boldsymbol{Q}^{(i)}),
    \label{eq:state_transition}
\end{equation}
where $i \in \mathcal{M} = \{ \text{CV}, \text{CA}, \text{CT+}, \text{CT-} \}$, $\boldsymbol{w}_t$ is the process noise vector in time step $t$, $\boldsymbol{F}^{(i)}$ and $ \boldsymbol{Q}^{(i)} $ are the state transition matrix and the process noise covariance matrix in the $ i $-th motion model, respectively.

\subsection{Measurement Model}
Consider a radar that accounts for tracking this maneuvering target. Its measurement $\boldsymbol{z}_t$ is composed of the target's position components along the three spatial axes, given by $\boldsymbol{z}_t = [l_x,\, l_y,\, l_z]^\top$. The measurement equation follows the linear Gaussian model given by 
\begin{equation}
\boldsymbol{z}_t = \boldsymbol{H} \boldsymbol{x}_t + \boldsymbol{v}_t,  \boldsymbol{v}_t \sim \mathcal{N}(\boldsymbol{0}, \boldsymbol{R}),
\end{equation}
where $\boldsymbol{H}$ is the measurement matrix, $\boldsymbol{v}_t$ is the measurement noise vector at time step $t$, and $\boldsymbol{R}$ is the measurement noise covariance matrix.

Based on the above dynamic model and measurement model, the objective of this paper is to design a MTT algorithm that combines both data-driven and model-driven methods to achieve the accurate tracking of this maneuvering target.

\section{The Proposed IMMNet Algorithm} \label{immnet}

\subsection{Algorithm Architecture}\label{3.2}
\begin{figure}[h]
	\centering
	\includegraphics[width=0.7\textwidth]{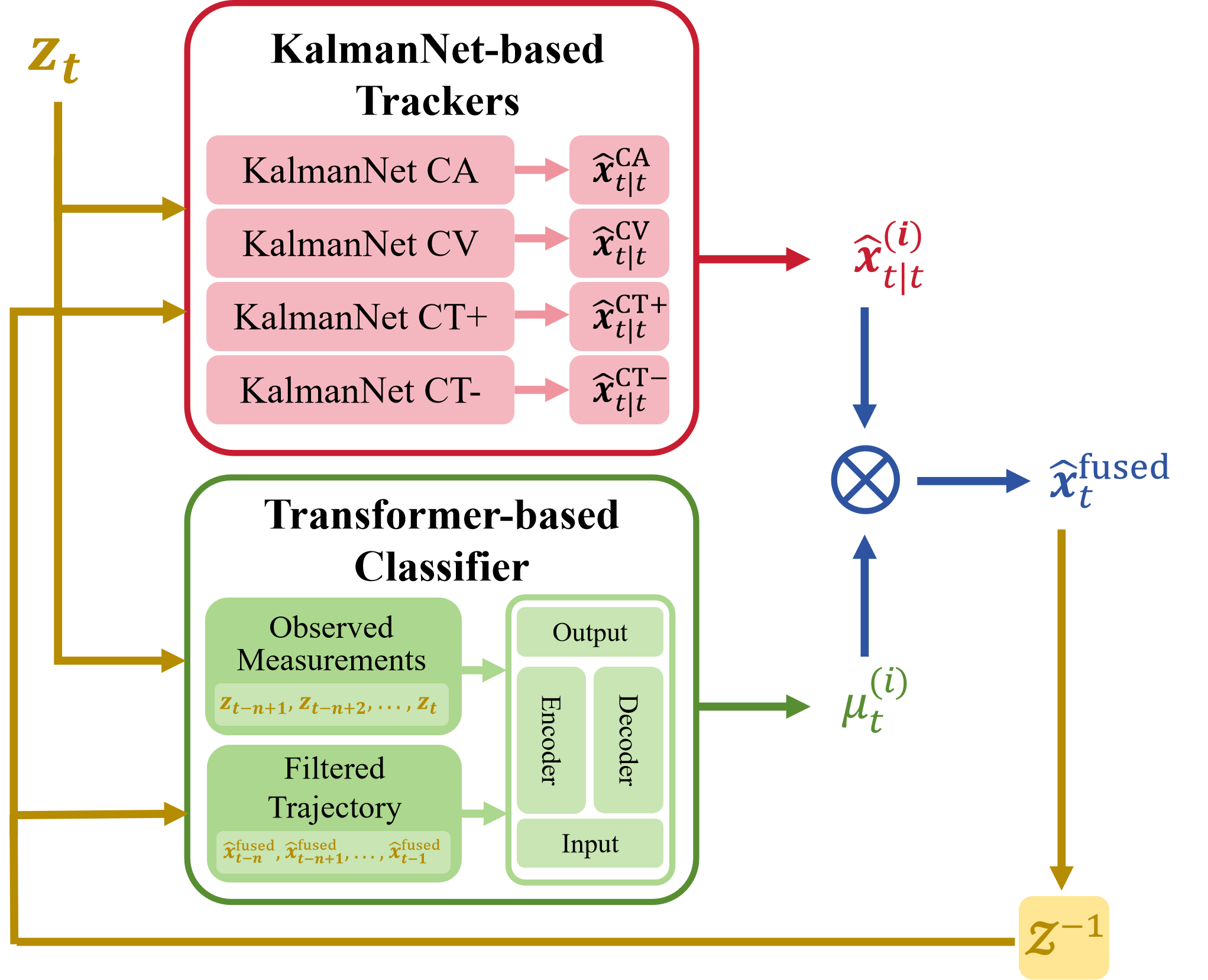}\\
	\caption{The Architecture of IMMNet.}\label{fig:immnet_architecture}
\end{figure}

Inspired by the existing IMM algorithm, we design the framework of the IMMNet algorithm by replacing each functional component of the IMM algorithm with a dedicated neural network module. Specifically, we substitute the conventional Kalman filters with KalmanNet-based trackers to perform state estimation, and we replace the fixed Markov-based model probability update mechanism with a Transformer-based classifier. The overall architecture of the proposed IMMNet algorithm is illustrated in Fig.~\ref{fig:immnet_architecture}.

The proposed IMMNet algorithm employs four parallel KalmanNet instances as expert trackers, each tailored to a distinct motion model: CV, CA, CT+, and CT-. These four models serve as predefined maneuver hypotheses, analogous to the model set in the IMM algorithm. Each KalmanNet is trained to specialize in filtering under its corresponding motion dynamics, thereby yielding more accurate state estimates when the target exhibits that specific behavior.

To replace the fixed Markov-based model probability update mechanism, we introduce a Transformer network with end-to-end trainable capability, which serves as a motion-mode classifier. At each time step $ t $, the input to this classifier includes two key components: the sequence of current and historical measurements, and the fused state output in the previous time step. The classifier then outputs a probability distribution across four distinct motion modes. This distribution effectively quantifies the likelihood that the target is performing each specific type of maneuver at time step $ t $.

The tracking procedure in the IMMNet algorithm consists of the following two sequential stages: parallel filtering and weighted fusion:

\begin{enumerate}
\item \textbf{Parallel Filtering}: At time step $ t $, each KalmanNet receives the same initial state estimate (typically the fused output from the previous time step) and processes it independently according to its designated motion model. Each tracker then produces a model-specific state estimate $\hat{\boldsymbol{x}}_t^{(i)}$ , where $ i $ indexes the motion mode.

\item \textbf{Weighted Fusion}: A temporal window of recent measurements and/or predicted states is fed into the Transformer-based classifier. Leveraging its attention mechanism, the classifier infers the current maneuvering mode by analyzing motion patterns over time and outputs a set of mode probabilities $ \mu_t^{(i)} $. These probabilities are then used as weights to fuse the individual KalmanNet estimates, i.e.,
\begin{equation}
    \hat{\boldsymbol{x}}_t^{\text{fused}} = \sum_{i\in \mathcal{M}}^{\mathcal{M}} \mu_t^{(i)} \hat{\boldsymbol{x}}_{t|t}^{(i)},
\end{equation}
where $ \mathcal{M} = \{\text{CV}, \text{CA}, \text{CT+}, \text{CT-}\} $ denote the set of motion modes. The fused state $ \hat{\boldsymbol{x}}_t^{\text{fused}} $ constitutes the final output of the IMMNet algorithm at time step $ t $.
\end{enumerate}

After establishing the above IMMNet architecture, three key issues remain to be addressed:
\begin{itemize}
\item The detailed architectural designs for both the KalmanNet trackers and the Transformer classifier need to be carefully investigated.
\item The appropriate features and data representations that serve as inputs to each component should be sufficiently explored.
\item Effective training strategies for the IMMNet algorithm, including both end-to-end and modular training paradigms, need to be studied.
\end{itemize}

These issues will be systematically addressed in the following subsections.

\subsection{Network Structure Design}\label{3.3}
\subsubsection{Tracker Design}
The primary role of each tracker in the IMMNet algorithm is to accurately estimate the target's state under its associated motion hypothesis while maintaining robustness against mild model mismatch. Unlike the IMM algorithm, whose performance degrades significantly under the case of model mismatch or unmodeled maneuvers, the IMMNet
algorithm can maintain tracking performance by replacing  the standard Kalman filter with KalmanNet.


Each KalmanNet is initialized according to its respective motion model’s state transition matrix. The input features to each KalmanNet follow the original KalmanNet formulation~\cite{revach2022kalmannet}, comprising:
\begin{itemize}
    \item F1: Observation difference $\Delta \tilde{\boldsymbol{z}}_t = \boldsymbol{z}_t - \boldsymbol{z}_{t-1}$,
    \item F2: Innovation difference $\Delta \boldsymbol{z}_t^{(i)} = \boldsymbol{z}_t - \hat{\boldsymbol{z}}_{t|t-1}^{(i)}$,
    \item F3: Forward evolution difference $\Delta \hat{\boldsymbol{x}}_{t} = \hat{\boldsymbol{x}}_{t|t} - \hat{\boldsymbol{x}}_{t-1|t-1}$,
    \item F4: Forward update difference $\Delta \hat{\boldsymbol{x}}_{t}^{(i)} = \hat{\boldsymbol{x}}_{t|t} - \hat{\boldsymbol{x}}_{t|t-1}^{(i)}$,
\end{itemize}
where $\boldsymbol{z}_t$ represents the measurement at time step $t$, $\hat{\boldsymbol{z}}_{t|t-1}^{(i)}$ is the predicted measurement of the $i$-th model, $\hat{\boldsymbol{x}}_{t|t}$ and $\hat{\boldsymbol{x}}_{t|t-1}^{(i)}$ denote the fused state estimate and predicted state estimate of the $i$-th model, respectively.
These features enable data-driven adaptation of the internal Kalman gain without explicit knowledge of process or measurement noise statistics.

\subsubsection{Classifier Design}
The inference process of the Transformer-based classifier in the IMMNet algorithm is to assign a probability weight to each motion mode according to the input feature, reflecting the likelihood that the target is currently executing that maneuver. Since each KalmanNet-based tracker performs optimally only under its designated dynamics, accurate mode classification is critical for effective fusion.

This inference process is inherently a sequential classification problem with strong temporal dependencies. As maneuvering targets exhibit frequent mode variations, motion mode information is primarily contained in short-term trajectories. Thus, the true motion mode is most discernible from short-term trajectory evolution. To enable robust inference, we feed the classifier with a sliding window of historical data that ends at the current time step $t$.

We propose two complementary feature streams as inputs:
\begin{itemize}
    \item F5: Observed measurements: Raw positional measurements over a time window of length $T$, i.e., $\{\boldsymbol{z}_{t-T+1}, \dots, \boldsymbol{z}_t\}$, which preserves real-world observation noise and sensor characteristics.
    \item F6: Filtered trajectory: The corresponding fused state estimates from the IMMNet algorithm over the same window, i.e., $\{\hat{\boldsymbol{x}}_{t-T+1}, \dots, \hat{\boldsymbol{x}}_t\} $, containing estimated positions, velocities, and accelerations.
\end{itemize}

\begin{figure}[h]
	\centering
	\includegraphics[width=0.5\textwidth]{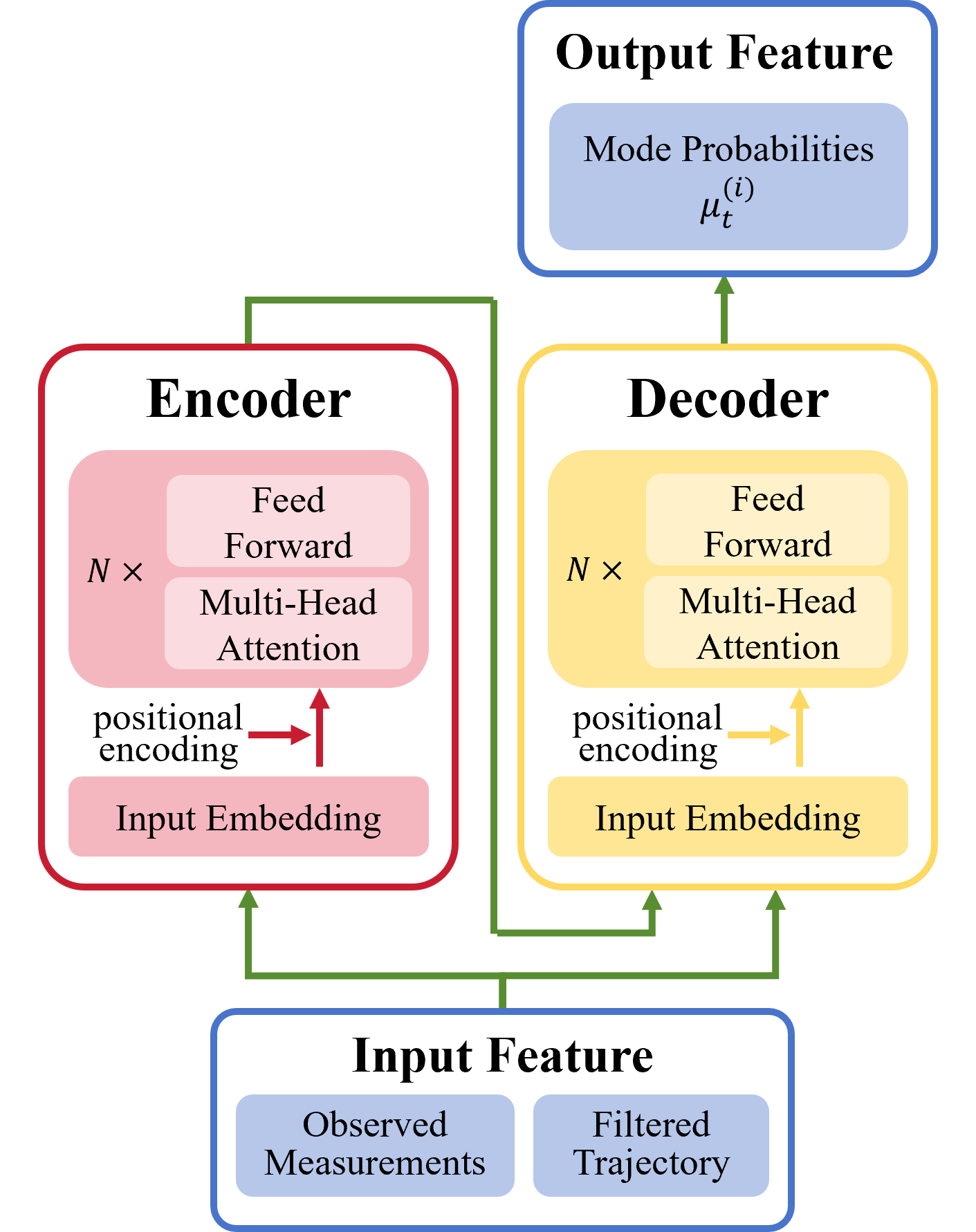}\\
	\caption{The architecture of the Transformer-based classifier.}\label{fig:classifier_arch}
\end{figure}

These two feature sequences are concatenated along the feature dimension and fed into a Transformer-based encoder, as shown in Fig.~\ref{fig:classifier_arch}. The self-attention mechanism captures intra-sequence dependencies within each modality, while cross-attention enables interaction between raw observations and refined state estimates. The final output is a 4-dimensional probability vector
\begin{equation}
    \boldsymbol{\mu}_t^{(i)} = [\mu_t^{\text{CV}},\, \mu_t^{\text{CA}},\, \mu_t^{\text{CT+}},\, \mu_t^{\text{CT-}}]^\top,
\end{equation}
representing the posterior probabilities of the four motion modes at time step $t$.

This inference process exhibits an autoregressive manner, which predicts the current motion mode by conditioning on a temporal context window. It enables the classifier to exploit both noisy  measurements and denoised dynamical cues. As a result, the proposed IMMNet algorithm can maintain high classification performance, even when the target performs transient maneuvers.

\subsection{Network Training}\label{3.4}
To simplify the training pipeline and enhance overall tracking accuracy, we adopt a two-stage distributed training strategy for the IMMNet algorithm.

\textbf{Stage 1: Tracker Pretraining.}  
We first train the four KalmanNet-based trackers independently. Leveraging KalmanNet’s ability to learn motion dynamics directly from data, each tracker is pretrained on a synthetic dataset containing both trajectories conforming to its designated motion model and general maneuvering target trajectories. To balance specialization and generalization, we set the ratio of model-consistent trajectories to arbitrary maneuvering trajectories as $6:4$. This encourages each KalmanNet tracker to excel under its native dynamics while retaining robustness to mild model mismatch.

The training objective is formulated as a supervised regression task, where the ground-truth state $\boldsymbol{x}_t$ is compared against the fused estimate from that individual tracker (denoted as $\hat{\boldsymbol{x}}_t^{\text{fused}}$, which in this stage equals its own output). According to~\cite{revach2022kalmannet}, the mean squared error (MSE) loss is adopted, i.e.,
\begin{equation}
    \mathcal{L}_{\text{Kal}} = \left\| \boldsymbol{x}_t - \hat{\boldsymbol{x}}_t^{\text{fused}} \right\|^2.
\end{equation}

\textbf{Stage 2: End-to-End Joint Training.}  
After pretraining, we assemble the full IMMNet architecture and perform end-to-end fine-tuning. In this stage, the primary focus shifts to optimize the classification accuracy of models, which directly governs fusion quality.

At each time step $t$, the ground-truth mode label is represented by a one-hot vector
\begin{equation*}
   \boldsymbol{C}_t=[C_t^{\text{CV}}, C_t^{\text{CA}}, C_t^{\text{CT+}}, C_t^{\text{CT-}}]^\top
\end{equation*}
 where $C_t^{(i)} \in \{0,1\}$,  $C_{t}^{(i)} = 1$ if the target is executing mode $i$ at time step $t$. The classifier outputs a probability distribution $\boldsymbol{\mu}_t = [\mu_{t}^{(i)}]_{i \in \mathcal{M}}$. We minimize the cross-entropy loss:
\begin{equation}
    \mathcal{L}_{\text{CE}} = -\sum_{i \in \mathcal{M}} C_{t}^{(i)}  \mu_{t}^{(i)}.
\end{equation}

The total loss combines tracking fidelity and classification correctness via a convex weighting scheme:
\begin{equation}
    \mathcal{L} = \gamma \, \mathcal{L}_{\text{Kal}} + (1 - \gamma) \, \mathcal{L}_{\text{CE}},
\end{equation}
where $\gamma \in [0,1]$ is a hyperparameter balancing the two objectives. In our experiments, we set $\gamma = 0.7$ to prioritize tracking accuracy while still enforcing meaningful mode discrimination.

The entire network is optimized using the Adam optimizer with a learning rate of $1 \times 10^{-3}$, batch size of 64, and trained for 100 epochs on the large-scale 3D maneuvering dataset described in Section~\ref{dataset}. Early stopping is applied based on validation loss to prevent overfitting.

\section{Large-Scale 3D Maneuvering Target Trajectory Dataset}\label{dataset}
To enhance the realism and applicability of the proposed IMMNet algorithm, we conduct all simulations and evaluations within a 3D Cartesian coordinate system. Accordingly, this paper constructs a large-scale dataset of 3D maneuvering target trajectories to facilitate effective training and comprehensive testing of the proposed IMMNet algorithm.
It is named 3D large-scale airborne surveillance trajectory (3D-LAST), in which each trajectory is explicitly designed to be maneuvering with multiple mode switches.

The 3D-LAST dataset is generated using the four motion modes introduced in Section~\ref{system}. The state-space model defined in Section~\ref{system} is employed. Crucially, at each time step $t$, a one-hot encoded label $C_t^{(i)} $ is assigned to indicate the active motion mode, enabling supervised training of the Transformer-based classifier.

\begin{figure*}[h]
	\centering
	\includegraphics[width=1\textwidth]{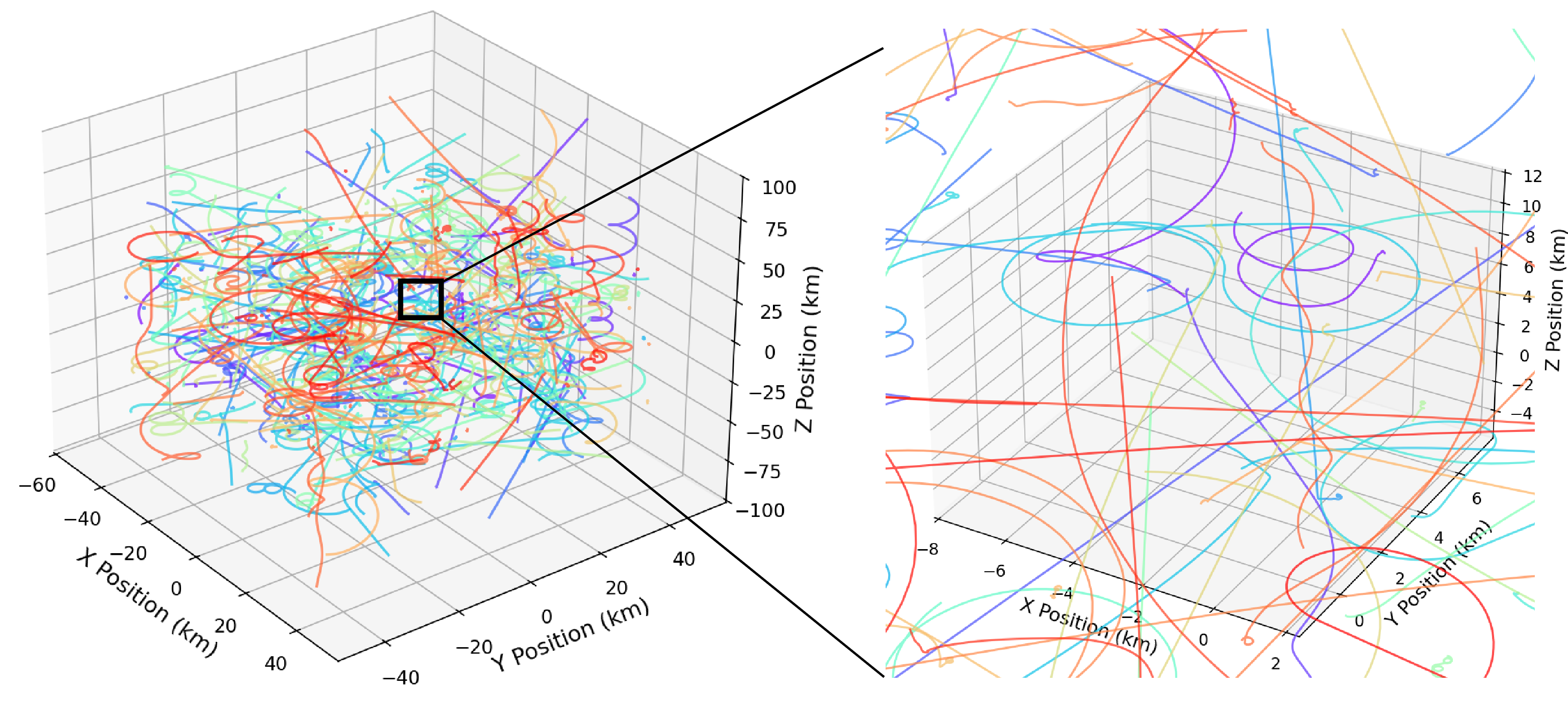}\\
	\caption{Overview of the 3D-LAST dataset, which displays 400 randomly sampled ground-truth trajectories.}\label{fig:dataset_overview}
\end{figure*}

Key generation parameters are summarized in Table~\ref{tab:dataset_params}. Specifically, the distance between radar and targets is constrained within $[0.5, 20]$ nautical miles (NM), i.e., approximately $[926, 37,\!040]$ meters, aligning with the effective coverage of conventional airport surveillance radar (ASR). Target speed is limited to $[0, 340]$ m/s, reflecting the subsonic regime typical of civil aviation. The turn rate $\alpha$ ranges from $-10^\circ/\text{s}$ to $+10^\circ/\text{s}$, consistent with standard aircraft maneuverability. Each trajectory spans $T = 400$ seconds.

\begin{table}[!t]
\renewcommand{\arraystretch}{1.2}
\caption{Parameters of the 3D-LAST Dataset}
\label{tab:dataset_params}
\centering
\begin{tabular}{lc}
\toprule
\textbf{Parameter} & \textbf{Value / Range} \\
\midrule
Distance to radar & $0.5$–$20$ NM ($926$–$37,\!040$ m) \\
Speed & $0$–$340$ m/s \\
Turn rate $\alpha$ & $-10^\circ/\text{s}$ to $+10^\circ/\text{s}$ \\
Total duration per trajectory & $400$ s \\
Number of trajectories & $100,\!000$ \\
Motion modes per trajectory & $3$ (randomly switched) \\
\bottomrule
\end{tabular}
\end{table}

To ensure physical plausibility and prevent trajectories from diverging beyond operational boundaries, we impose soft constraints during generation: When a trajectory approaches the spatial or kinematic limits, its motion mode is automatically switched to a CT model, which naturally confines the target within a bounded region. This strategy not only avoids out-of-bound trajectories but also guarantees uniform sequence length across the entire dataset.
Within each generated trajectory, two random time instants are selected as mode-switching points. At these time instants, the motion model transitions to a different mode, with the constraint that no mode is repeated consecutively. This design ensures rich intra-trajectory dynamics while maintaining diversity across the dataset.

\begin{figure}[h]
	\centering
	\includegraphics[width=0.6\textwidth]{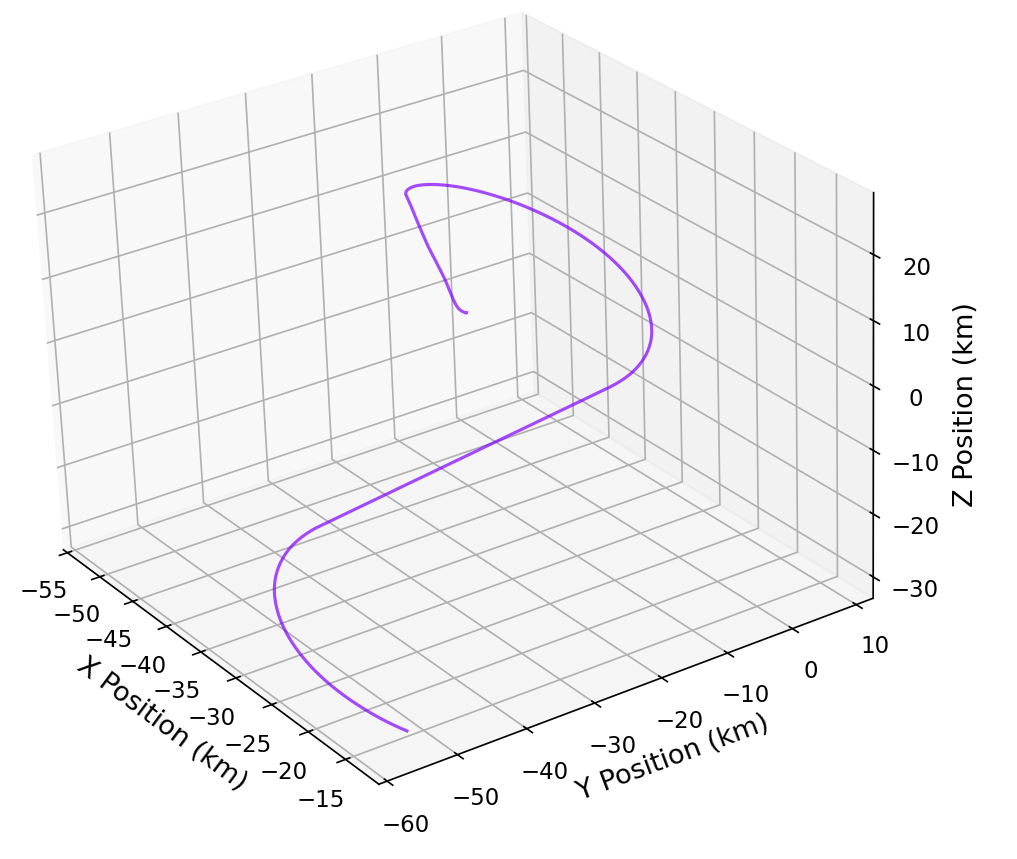}\\
	\caption{A representative trajectory in the 3D-LAST dataset.}\label{fig:tarrandom}
\end{figure}

The final dataset consists of two synchronized components for each trajectory: The ground-truth state sequence and the corresponding noisy observation sequence. 
In total, 100,000 such trajectories are generated.
Fig.~\ref{fig:dataset_overview} displays 400 randomly sampled ground-truth trajectories, demonstrating extensive coverage of the 3D airspace and a wide spectrum of maneuvering behaviors.
Fig.~\ref{fig:tarrandom} illustrates a random trajectory from the 3D-LAST. The target first performs a CA motion to change its direction, then performs a CV motion, and finally switches to a CT motion for turning. Such multi-phase maneuvers with explicit mode transitions constitute the core characteristic of this dataset.

In summary, the proposed 3D-LAST dataset features a complete 3D state description, explicit multi-mode-switching within trajectories and physically reasonable boundary constraints. It provides a high-quality, large-scale, and reproducible benchmark for developing and evaluating 3D maneuvering target tracking algorithms. The open-source resources of this dataset are available\footnote{https://github.com/STFLAB/3D-LAST-dataset}.


\section{Experiments} \label{EXPERIMENTAL}
In this section, we evaluate the tracking performance of  the proposed IMMNet algorithm over the 3D-LAST dataset.

\subsection{Experimental Setup}
\label{subsec:setup}

The 3D-LAST dataset is generated with measurement noise standard deviation $r$ uniformly sampled from $[1, 7]$\,m and process noise standard deviation $\sigma_q = 0.2$\,m/s$^2$. All neural models are implemented in PyTorch 1.11 and trained on an NVIDIA GeForce RTX 4090 GPU. The detailed training configurations for both KalmanNet and the proposed IMMNet algorithm are summarized in Table~\ref{tab:training_params}.

\begin{table}[!t]
\centering
\caption{Training Hyperparameters}
\label{tab:training_params}
\begin{tabular}{lcc}
\toprule
\textbf{Parameter} & \textbf{KalmanNet} & \textbf{IMMNet} \\
\midrule
Training epochs     & 300 & 20 \\
Learning rate       & 0.01 & 0.001 \\
Batch size          & 100 & 100 \\
Optimizer           & Adam & Adam \\
Window length       & 20 & 20 \\
\bottomrule
\end{tabular}
\end{table}

It is worth noting that the measurement noise standard deviation $r$ and process noise standard deviation $\sigma_q$ are unknown for the proposed IMMNet algorithm, while IMM algorithm can only work properly when these two parameters are known. Thus, three IMM variants are designed for comprehensive performance evaluation, including IMM-Accurate with fully known parameters, IMM-Max with only the upper bound of measurement noise standard deviation $r$, and IMM-Min with only the lower bound of measurement noise standard deviation $r$. IMM-Accurate adopts measurement noise standard deviation $r=4$ and process noise standard deviation $\sigma_q=0.2$. IMM-Max adopts measurement noise standard deviation $r=7$ and process noise standard deviation $\sigma_q=1$. IMM-Min adopts measurement noise standard deviation $r=1$ and process noise standard deviation $\sigma_q=1$. Three different IMM algorithm variants are initialized with CV, CA and CT motion models and the initial model probabilities are set to $[1/3, 1/3, 1/3]$.
The mode transition matrix is defined as
\begin{equation}
\boldsymbol{M} = 
\begin{bmatrix}
0.8 & 0.1 & 0.1 \\
0.1 & 0.8 & 0.1 \\
0.1 & 0.1 & 0.8
\end{bmatrix}.
\label{eq:mode_transition}
\end{equation}
The initial state estimate for the IMM algorithms is set to the first observation in each trajectory.

After training, both the IMMNet and IMM algorithms are evaluated on a test set. The evaluation proceeds in two stages:  
(1) Global performance assessment, we compute the average root mean square error (RMSE) across all trajectories in the test set to quantify the overall estimation accuracy.  
(2) Case study analysis, we select several random maneuvering trajectories and conduct a fine-grained comparison. 

\subsection{Performance Evaluation}
\label{subsec:overall_performance}

\begin{figure*}[!t]
    \centering
    \subfigure[RMSE of trajectory 1.]{\label{fig:line1}
    \includegraphics[width=0.9\textwidth]{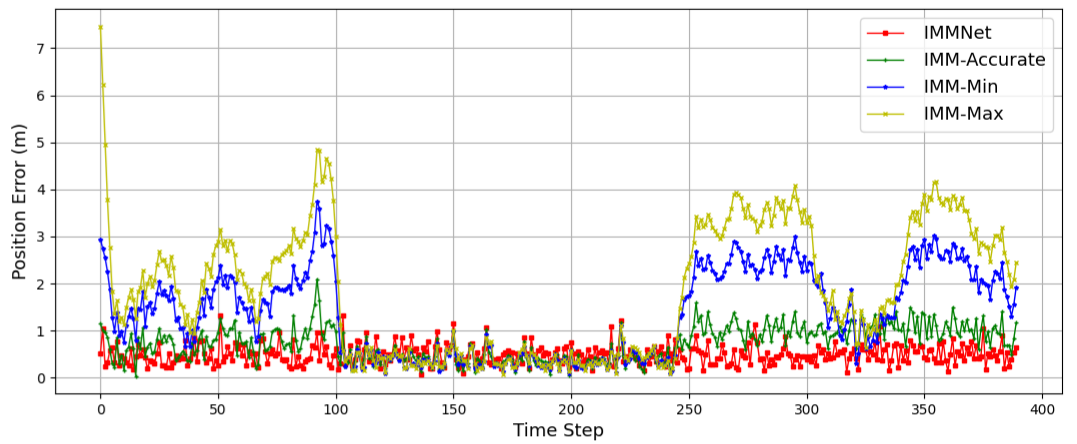}}
    \par
    \subfigure[Tracking results on trajectory 1 with enlarged view near mode-switching point.]{\label{fig:no1}
    \includegraphics[width=0.9\textwidth]{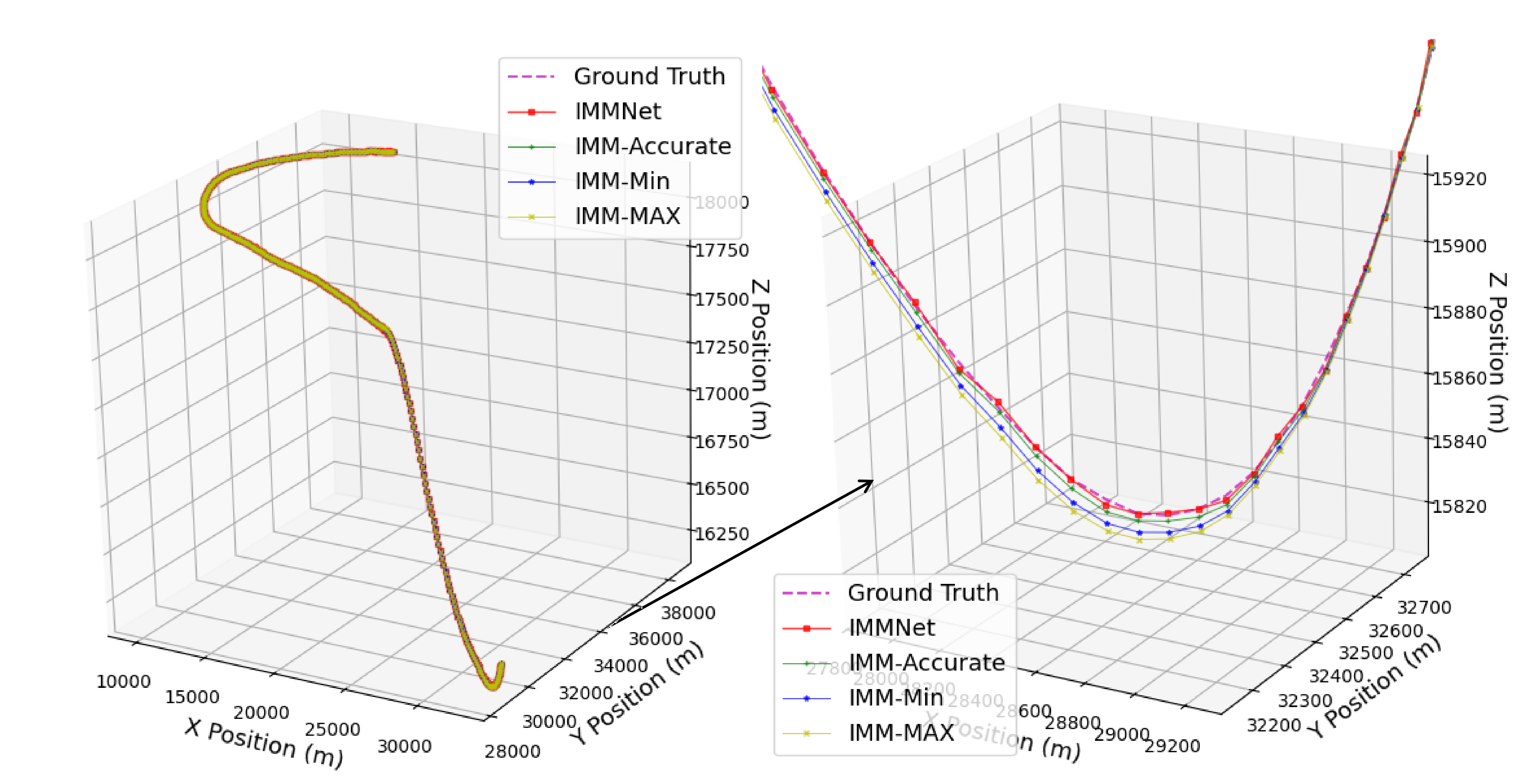}}
    \caption{Experimental result of trajectory 1.}
    \label{fig:total1}
\end{figure*}

Table~\ref{tab:overall_rmse} reports the RMSE of both the IMMNet and the IMM algorithms over the entire test set. 
As one can see, the IMMNet algorithm achieves significantly lower RMSE than all IMM algorithm variants across all spatial dimensions, with an average error of 0.29 m compared to 1.32 m, 3.29 m and 4.55 m. This demonstrates that, the proposed IMMNet algorithm effectively learns both the kinematic patterns of maneuvering targets and the statistical characteristics of trajectory noise, enabling more accurate state estimation and superior noise suppression during filtering.

\begin{table}[!t]
\centering
\caption{Overall average RMSE over the Full Test Set (m)}
\label{tab:overall_rmse}
\begin{tabular}{lcccc}
\toprule
Algorithm   & $x$-axis & $y$-axis & $z$-axis & Average \\
\midrule
IMMNet   & 0.29 & 0.29 & 0.29 & \textbf{0.29} \\
IMM-Accurate      & 1.60 & 1.63 & 0.72 & \textbf{1.32} \\
IMM-Min      & 4.13 & 3.91 & 1.83 & \textbf{3.29} \\
IMM-Max      & 5.82 & 5.21 & 2.62 & \textbf{4.55} \\
\bottomrule
\end{tabular}
\end{table}

We further analyze the tracking performance on two random trajectories exhibiting distinct maneuvering behaviors.

\textbf{Trajectory 1:} As illustrated in Fig.~\ref{fig:total1}, the test results of trajectory 1 are presented. Fig.~\ref{fig:line1} shows the time-varying RMSE curves and Fig.~\ref{fig:no1} presents the tracking performance along the trajectory with enlarged views near mode-switching points. It can be observed from Fig.~\ref{fig:line1} that the proposed IMMNet algorithm achieves the best tracking performance throughout the trajectory. Large errors arise for the IMM algorithm after mode-switching due to model mismatch. As a contrast, the proposed IMMNet algorithm maintains stable performance during all periods. It can also be seen from Fig.~\ref{fig:no1} that the IMMNet algorithm provides satisfactory tracking results. The enlarged views verify that the proposed IMMNet algorithm yields the smallest error around mode-switching points. The RMSE values are shown in Table~\ref{tab:traj1_rmse}. These results confirm that the IMMNet algorithm provides consistently better tracking accuracy throughout the trajectory, especially after the motion mode switches.

\begin{table}[!h]
\centering
\caption{RMSE for trajectory 1 (m)}
\label{tab:traj1_rmse}
\begin{tabular}{lcccc}
\toprule
Algorithm   & $x$-axis & $y$-axis & $z$-axis & Average \\
\midrule
IMMNet   & 0.47 & 0.50 & 0.52 & \textbf{0.50} \\
IMM-Accurate      & 0.79 & 0.83 & 0.57 & \textbf{0.73} \\
IMM-Min      & 1.46 & 1.75 & 0.90 & \textbf{1.37} \\
IMM-Max      & 1.95 & 2.39 & 1.17 & \textbf{1.84} \\
\bottomrule
\end{tabular}
\end{table}

\begin{figure*}[!t]
    \centering
    \subfigure[RMSE of trajectory 2.]{\label{fig:line2}
    \includegraphics[width=0.9\textwidth]{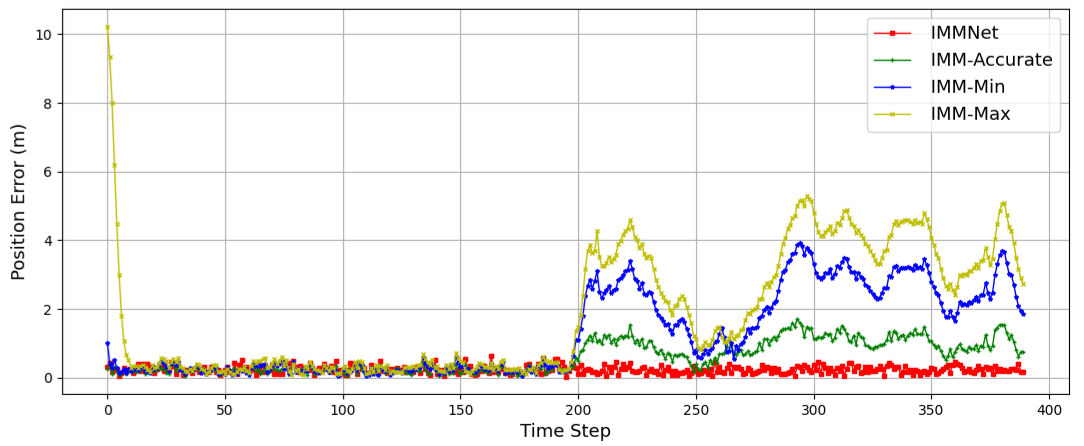}}
    \par
    \subfigure[Tracking results on trajectory 2 with enlarged view near mode-switching point.]{\label{fig:no2}
    \includegraphics[width=0.9\textwidth]{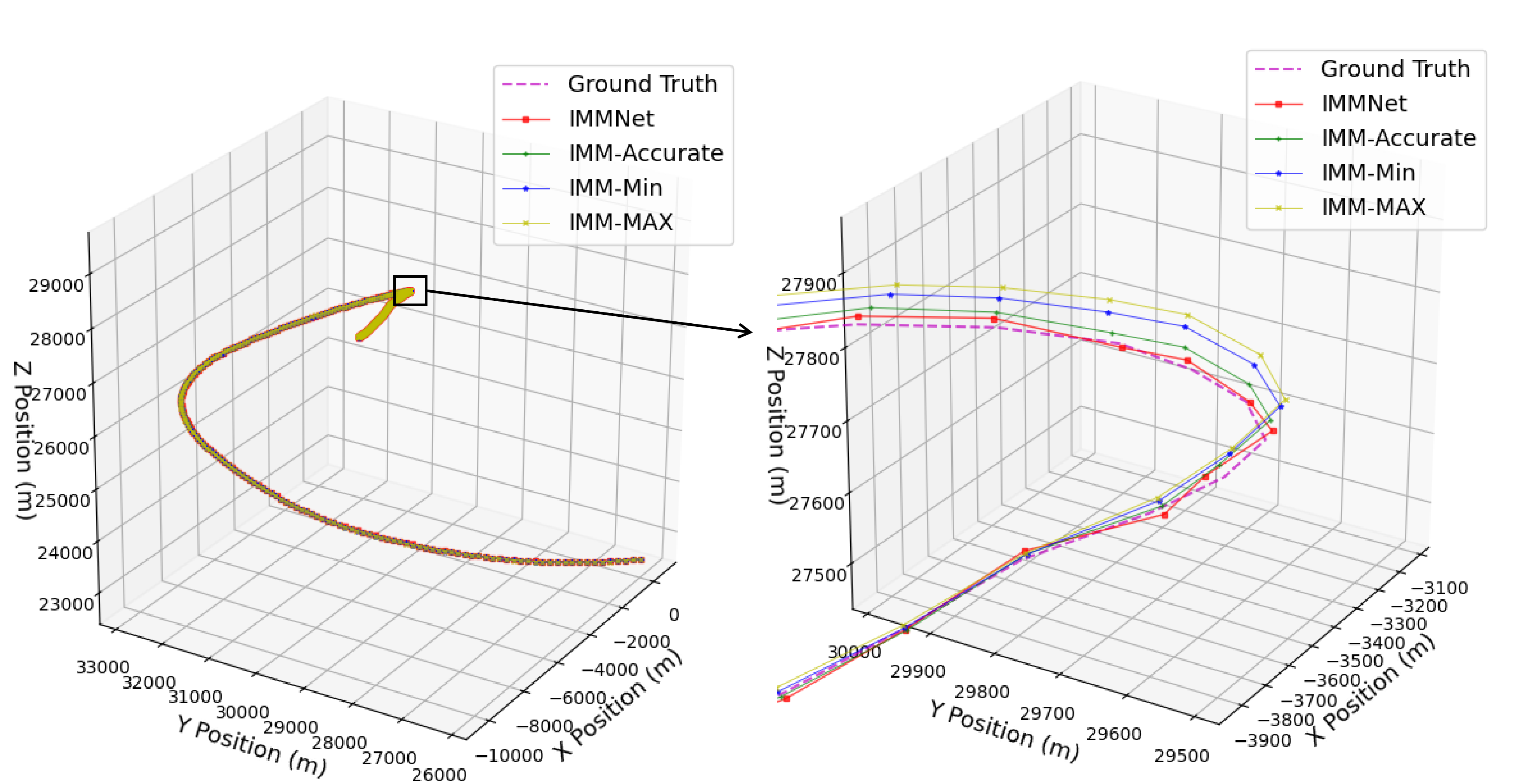}}
    \caption{Experimental result of trajectory 2.}
    \label{fig:total2}
\end{figure*}

\textbf{Trajectory 2:} As shown in Fig.~\ref{fig:total2}, the tracking results of trajectory 2 are presented. Fig.~\ref{fig:line2} shows the time-varying RMSE curves and Fig.~\ref{fig:no2} displays the tracking performance along the trajectory with enlarged view near mode-switching point. Similar to the results of trajectory 1, the IMM algorithm arises large fluctuations after mode-switching point while the IMMNet algorithm remains stable. The proposed IMMNet algorithm also achieves superior performance near the mode-switching point as shown in  Fig.~\ref{fig:no2}. Similarly, the IMMNet algorithm maintains lower overall error. The RMSE values are shown in Table~\ref{tab:traj2_rmse}.

\begin{table}[!h]
\centering
\caption{Average RMSE for trajectory 2 (m)}
\label{tab:traj2_rmse}
\begin{tabular}{lcccc}
\toprule
Algorithm   & $x$-axis & $y$-axis & $z$-axis & Average \\
\midrule
IMMNet   & 0.24 & 0.23 & 0.24 & \textbf{0.24} \\
IMM-Accurate      & 0.45 & 0.84 & 0.50 & \textbf{0.59} \\
IMM-Min      & 0.84 & 1.95 & 1.10 & \textbf{1.30} \\
IMM-Max      & 1.19 & 2.80 & 1.69 & \textbf{1.89} \\
\bottomrule
\end{tabular}
\end{table}

In summary, experimental results validate that the IMMNet algorithm significantly outperforms the IMM algorithm in tracking maneuvering targets across diverse motion patterns. On the full test set, it achieves an lower average RMSE than the IMM algorithm and consistently demonstrates lower estimation error in both steady-state and mode-switching point phases of representative trajectories. This confirms that the proposed IMMNet algorithm effectively learns target dynamics and noise characteristics, enabling more accurate and adaptive state estimation.

\section{Conclusion} \label{conclusion}
This paper proposed the IMMNet algorithm, a hybrid model/data-driven algorithm that synergistically combines the structural interpretability of the IMM algorithm with the adaptive learning capability of neural networks. Extensive experiments on a large-scale 3D maneuvering target dataset demonstrate that the IMMNet algorithm consistently outperforms the IMM algorithm in terms of average RMSE across diverse motion patterns. These results validate the effectiveness of embedding learnable modules within a principled multi-model architecture to mitigate model mismatch and improve tracking performance. In the future, we will focus on enhancing the noise robustness of the IMMNet algorithm and extending the IMMNet algorithm to multi-target scenarios  for more practical and complex tracking applications.

\bibliographystyle{elsarticle-num}        
\bibliography{new}    

%
%
%

\end{document}